\documentclass{article}

 \usepackage[dblblindworkshop, final]{neurips_2025}
\title{Diffusion for Fusion: Designing Stellarators with Generative AI}
\workshoptitle{AI4Science}

\usepackage{macros}

\usepackage[utf8]{inputenc} % allow utf-8 input
\usepackage[T1]{fontenc}    % use 8-bit T1 fonts
\usepackage{hyperref}       % hyperlinks
\usepackage{url}            % simple URL typesetting
\usepackage{booktabs}       % professional-quality tables
\usepackage{amsfonts}       % blackboard math symbols
\usepackage{nicefrac}       % compact symbols for 1/2, etc.
\usepackage{microtype}      % microtypography
\usepackage{xcolor}         % colors

\usepackage{duckuments}
\usepackage{hyperref}
\usepackage[normalem]{ulem}

\author{%
  Misha Padidar \orcidlink{0000-0002-0710-4377}, Teresa Huang \orcidlink{0000-0001-6631-3737},
  Andrew Giuliani \orcidlink{0000-0002-4388-2782},
  Marina Spivak \\
  Center for Computational Mathematics\\
  Flatiron Institute\\
  New York, NY 10010 \\
  \texttt{(mpadidar, thuang, agiuliani, mspivak)@flatironintsitute.org} \\
}

\begin{document}

\maketitle

\begin{abstract}
  Stellarators are a prospective class of fusion-based power plants that confine a hot plasma with three-dimensional magnetic fields. 
  Typically framed as a PDE-constrained optimization problem, stellarator design is a time-consuming process that can take hours to solve on a computing cluster. Developing fast methods for designing stellarators is crucial for advancing fusion research. Given the recent development of large datasets of optimized stellarators, machine learning approaches have emerged as a potential candidate. Motivated by this, we present an \textit{open} inverse problem to the machine learning community: to rapidly generate high-quality stellarator designs which have a set of desirable characteristics. As a case study in the problem space, we train a conditional diffusion model on data from the QUASR database to generate quasisymmetric stellarator designs with desirable characteristics (aspect ratio and mean rotational transform). The diffusion model is applied to design stellarators with characteristics not seen during training. We provide evaluation protocols and show that many of the generated stellarators exhibit solid performance: less than $5\%$ deviation from quasisymmetry and the target characteristics. The modest deviation from quasisymmetry highlights an opportunity to reach the sub $1\%$ target.
  Beyond the case study, we share multiple promising avenues for generative modeling to advance stellarator design.
\end{abstract}

\section{Introduction}
\label{sec:introduction}
Stellarators are a prospective class of  power plants that generate renewable energy with nuclear fusion \cite{imbert-gerard_introduction_2024}. The toroidal devices confine a hot plasma using strong magnetic fields generated by electromagnetic coils. Unlike their axisymmetric cousin, the tokamak, stellarators have magnetic fields with full three-dimensional shaping. Due to the complexity of the magnetic fields, stellarator design is typically framed as a PDE-constrained shape optimization problem, which requires running computationally intensive numerical solvers and evaluators \cite{hegna_infinity_2025,bonofiglo_fast_2025,carbajal_alpha-particle_2025,najmabadi_aries-cs_2008}.
%the design of stellarators is performed algorithmically, taking the form of a PDE-constrained shape optimization problem, which often runs for hours on a computing cluster. After optimization, computationally intensive simulations are run to quantify a device's performance at high fidelity \cite{hegna2025infinity,bonofiglo2025fast,carbajal2025alpha,najmabadi2008aries}. Altogether, designing stellarators is a computationally and time intensive problem requiring hours, days, or even weeks on a computing cluster. 
To advance research in fusion energy, it is crucial to accelerate stellarator design. Leveraging recent advances in simulation capabilities, the stellarator community has opened the doors to \textit{data-driven} stellarator design by compiling large datasets of optimized stellarators \cite{giuliani_comprehensive_2025,giuliani_direct_2024,cadena_constellaration_2025,landreman_mapping_2022,curvo_using_2025,gaur_omnigenous_2024}. 
% The datasets can be used to uncover hidden symmetries and relationships \cite{landreman2025does}, to aggregate coarse analytic models and sparse high-fidelity data into surrogates for quantities that are expensive to simulate, or to rapidly generate stellarators with promising performance. 
Using these datasets, machine learning models can uncover hidden relationships  \cite{landreman_how_2025,sengupta_periodic_2025}, and rapidly generate stellarators with promising performance \cite{curvo_using_2025}. 
% Through leveraging existing data and encoding well known problem symmetries and structure, \blue{generative machine learning methods have emerged as a promising approach for scientific applications, such as molecular design \cite{gomez2018automatic,jumper2021highly} and material design \cite{merchant2023scaling}. Generative models are well positioned to have a significant impact on stellarator design, capable of uncovering hidden relationships in the data and generating candidate designs rapidly. }
Given the past impact of generative machine learning methods on scientific applications, such as molecular design \cite{gomez-bombarelli_automatic_2018,jumper_highly_2021}, material design \cite{merchant_scaling_2023,choudhary_recent_2022}, and astrophysics \cite{mudur_can_2022,moriwaki_deep_2020,karchev_strong-lensing_2022,zhao_can_2023} generative models are well positioned to have an impact on stellarator design. 
% Through leveraging existing data and encoding well known problem symmetries and structure generative models are capable of uncovering hidden relationships in the data and generating candidate designs rapidly.

The aim of this paper is to state an \textit{open inverse problem} in stellarator design as a data-driven, generative modeling task, and to provide an entry point for the generative modeling community to advance stellarator research. The problem of interest is as follows,
\begin{namedproblem}[\bf{(IP)}]
    Given an array of desirable stellarator characteristics $\yb \in\R^{\ny}$, generate a stellarator design which has properties $\yb$. 
    \label{prob:main}
\end{namedproblem}
\vspace{-5pt}
Rapidly solving \nameref{prob:main} is the central goal of stellarator design. The main challenge being that the inverse mapping from properties, $\yb$, to stellarator designs is not directly available. 
% In fact, while the mapping, $\Fb:\R^{\nx}\to\R^{\ny}$, from design variables, $\xb$, to stellarator properties, $\yb$, is well defined, it may not be injective, i.e. distinct stellarators map to the same design property, i.e., $\Fb(\xb_1)=\Fb(\xb_2)=\yb$ and the inverse mapping $\Fb^{-1}(\yb)$ is multi-valued. 
% \mpnote{@Andrew: how about \say{In fact, while the mapping, $\Fb:\R^{\nx}\to\R^{\ny}$, from design variables, $\xb$, to stellarator properties, $\yb$, is well defined, it may not be invertible since many distinct designs have the same properties.}}
In fact, while the function, $\Fb:\R^{\nx}\to\R^{\ny}$, mapping design variables, $\xb$, to stellarator properties, $\yb$, is well defined, it may not be invertible since many distinct designs have the same properties.
As a result, numerical optimization is typically used to vary $\xb$ until $\Fb(\xb)$ is close to a desired $\yb$. 
Generative models, however, open the doors to a new probabilistic approach to this problem. Generative models can \say{invert} $\Fb$ in a probabilistic sense, by finding a density with support over the set of $\xb$ that map to $\yb$. In probabilistic terms, the generative model would learn a sampling mechanism (or conditional probability density) to produce features $\xb$ from conditions $\yb$, $\xb \sim p(\xb |\yb)$, such that $\Fb(\xb) \approx \yb$. 

To facilitate working with stellarator data, we provide an introduction to stellarator design, which discusses design variables $\xb$, properties of stellarators $\yb$, methods of evaluation $\Fb$, and current approaches to designing stellarators (see \cref{sec:stellarator_review}). In addition, we provide a concrete case study, in which we train denoising diffusion models \cite{ho_denoising_2020} to generate stellarator designs with $\nx=661$ features given $\ny=4$ conditions, using data drawn from the QUASR database (see \cref{sec:case_study}) \cite{giuliani_comprehensive_2025, giuliani_direct_2024}. 
% We quantify the performance of the stellarators generated by the diffusion model by evaluating them with \emph{two evaluation protocols:} the non-differentiable Ideal Magnetohydrodynamic (MHD) code, \texttt{VMEC} \blue{CITE}, and \blue{the differentiable evaluator to assess the generated design from the downstream coil configuration} \tes{A name for this evaluator?}. \mpnote{Discuss results and state that we discuss directions for improvement.}
The diffusion model is applied to design stellarators with characteristics not seen during training. Many of the generated stellarators exhibit solid performance: a low deviation from quasisymmetry and good adherence to the conditions.
% We find that the model gives satisfactory performance at generating devices given conditions from the datset, and that the model can successfully generate stellarators with properties never before seen. 
% This case study acts both as an introduction to the problem space, and as a baseline for model performance. 
To open the doors to further research, we discuss several promising research directions, from physics informed loss design, to dataset curation, model architecture design, sampling procedures, and multi-fidelity models (see \cref{sec:open_problems}).

\section{Stellarator Design: Preliminaries, Problem Set-up, and Related Work}
\label{sec:stellarator_review}

\begin{figure}[tbh!]
    \centering
    \resizebox{\textwidth}{!}{
    \begin{tikzpicture}

        \node (A) at (-0.75,0) {\includegraphics[height=0.175\textheight]{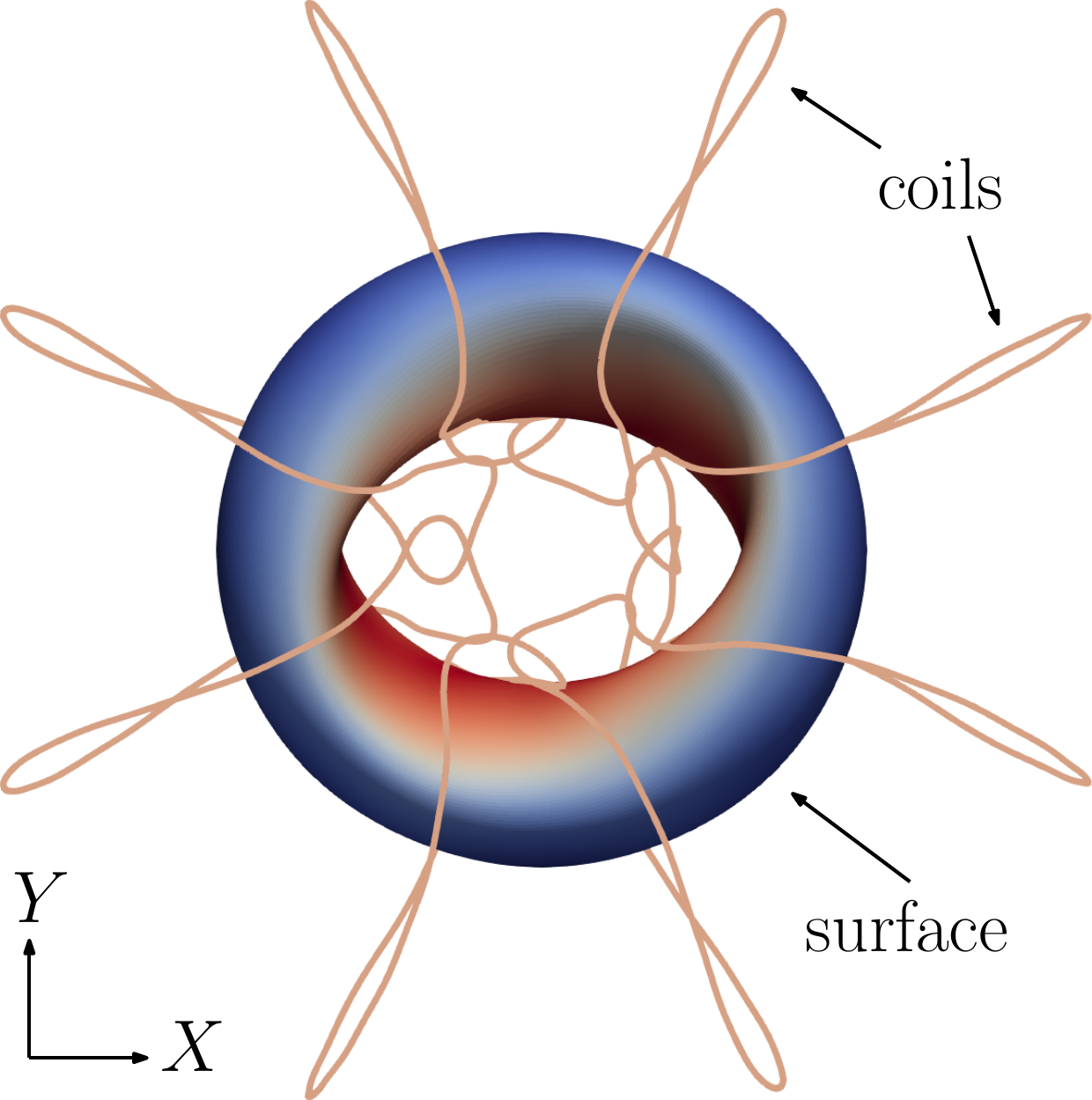}};
        \node (B) at (4.5,0) {\includegraphics[height=0.175\textheight]{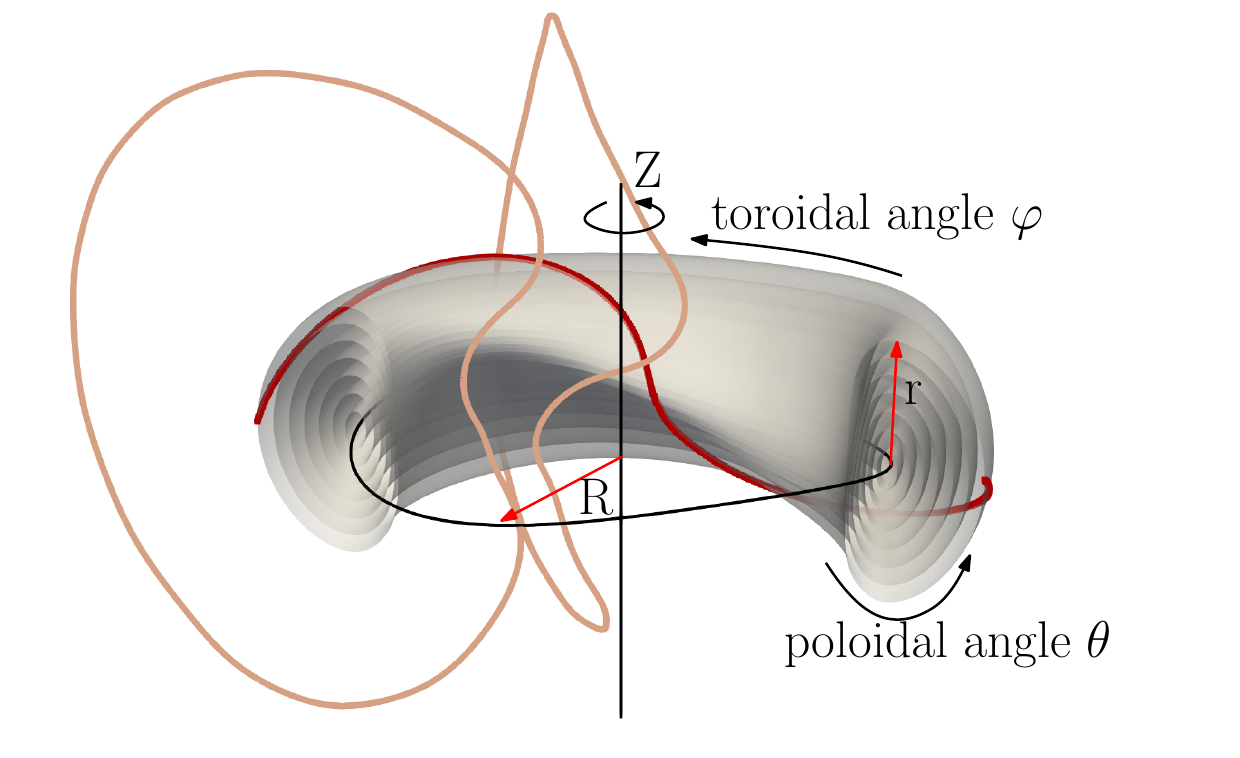}};
        \node (C) at (9.1,0) {\includegraphics[height=0.175\textheight]{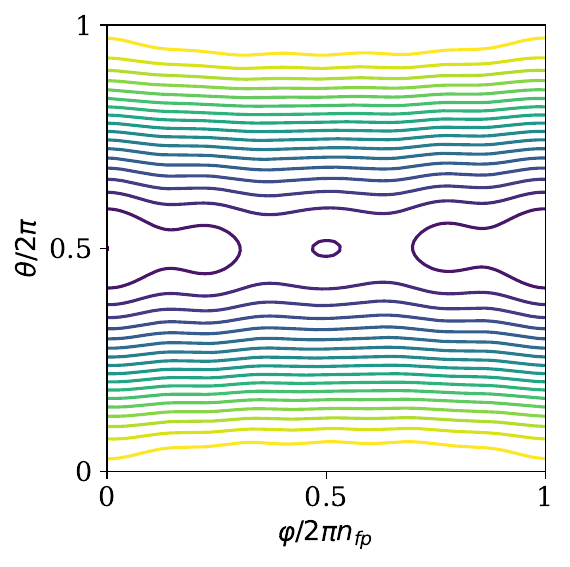}};
        
        \node (Alabel) at (-1.95, 1.6) {\large A};
        \node (Blabel) at (3.65, 1.6) {\large B};
        \node (Clabel) at (8.25, 1.6) {\large C};
    \end{tikzpicture}
    }
    \caption{(A) Stellarator-\texttt{0010198} from the QUASR database, encircled by the modular electromagnetic coils (gold) that generate its magnetic field (more information about this device can be found at \href{https://quasr.flatironinstitute.org/model/0010198}{[link]}). The surface is colored by the strength of the magnetic field. (B) A slice of the device in (A), with internal magnetic surfaces plotted. The device's minor radius, $r$, and major radius, $R$, are illustrated along with the directions in which the toroidal and poloidal angles increase on the magnetic surfaces. The device presents $n_{\text{fp}}=2$ discrete rotational symmetry about the $Z$ axis.  The red curve indicates a hypothetical field line. (C) isolines of field strength $\| \mathbf B \|$ on a magnetic surface, where $\varphi, \theta$ are the toroidal and poloidal Boozer angles on the surface. }
    \label{fig:stellarator_example}
\end{figure}

Stellarators use non-axisymmetric magnetic fields, $\Bb$, to confine plasma, a hot soup of ionized gas, to torus-like volumes \cite{imbert-gerard_introduction_2024}. These fields can be generated using electromagnetic coils (Figure \ref{fig:stellarator_example}A) and often form a region of nested magnetic surfaces (Figure \ref{fig:stellarator_example}B), which are everywhere tangent the magnetic field.
A stellarator's magnetic field satisfies the ideal MHD equations
\begin{equation}
\begin{aligned}\label{eq:idealMHD}
    \mu_0^{-1}(\nabla\times\Bb)\times\Bb &= \nabla p,
    \\
    \nabla \cdot \Bb &= 0,
    \\
    \Bb\cdot \nb &= 0 \quad \text{on} \, \, S(\xb),
\end{aligned}
\end{equation}
where $S$ is the shape of the plasma boundary (a magnetic surface), $p$ is the pressure profile, $\mu_0$ is the permeability of free space, and $\nb$ is the surface normal. 
The geometry of $S$ is given by the design variables $\xb$, which are typically Fourier harmonics (see \Cref{sec:surface_description} for more information).
In this work we only consider \textit{vacuum} magnetic fields, those with zero current, $\nabla\times\Bb = 0$ -- a common simplification. Given the shape of the plasma boundary $S$, the ideal MHD equations can be solved to obtain $\Bb$ with the \texttt{VMEC}, \texttt{DESC}, \texttt{GVEC} or \texttt{SPEC} codes \cite{hirshman_preconditioned_1991,dudt_desc_2020,hindenlang_computing_2025,hudson_computation_2012}; \texttt{VMEC} has long been the workhorse of the stellarator design community, but, unlike \texttt{DESC}, does not give access to derivatives with respect to design parameters.

Stellarator design is focused, in part, on manipulating properties of magnetic fields with nested surfaces. 
% The magnetic field vectors lay tangent to the surfaces, which can be described by the boundary condition $\Bb\cdot \nb = 0 $. 
Points on a magnetic surface can be described with two angles: the poloidal angle, $\theta$, and toroidal angle, $\varphi$, see \Cref{fig:stellarator_example}B; the toroidal angle varies the long way around the torus while the poloidal angle varies the short way around the torus. A magnetic surface has three basic attributes: the number of field periods $\nfp$, the aspect ratio $A$, and the rotational transform, $\iota$. The number of field periods, $n_{\text{fp}}$, refers to a discrete rotational symmetry, i.e., rotating the device by $2\pi/n_{\text{fp}}$ about the $Z$-axis results in the same device. A device with $\nfp$-field periods can be sliced (like a pizza) into $\nfp$ identical pieces; for example, \cref{fig:contours_in_sample} (top row) shows devices with $\nfp=2,3,4,5,6,7,8$, from left to right. The aspect ratio, $A = R/r$, is a geometric quantity measuring the ratio of the major to minor radius of the device, see \Cref{fig:stellarator_example}B, and reflects how thin the enclosed plasma is. Low aspect ratio devices are often preferred, since they correspond to a high volume of plasma confined, relative to the footprint. 
% A surface can also be described through its major and minor radii, $R, r$, respectively: the minor radius describing the average minor radius of a contant-$\varphi$ cross-section, and the major radius capturing the average cylindrical radius coordinate, $\sqrt{x^2(\theta,\varphi) + y^2(\theta,\varphi)}$. 
A third property which characterizes magnetic surfaces is the rotational transform $\iota$. The rotational transform is the average number of poloidal rotations a field line makes for every toroidal turn around the torus. \Cref{fig:stellarator_example}B illustrates a hypothetical red field line making, roughly, one poloidal turn as it transits one field period, leading to $\iota\approx \nfp=2$. When given multiple nested magnetic surfaces with different values of $\iota$, the average value, $\meaniota$, is considered.
% There is a complex interplay between rotational transform and quasisymmetry that has implications on quality of particle confinement.
When designing stellarators, the shape of $S$ is varied to achieve desirable values of $\meaniota, A$. 
% The shape of $S$ can be parameterized by a feature vector $\xb$, which often is a set of Fourier coefficients for two-dimensional Fourier series representing the coordinates of $S$, $[x(\varphi, \theta), y(\varphi, \theta), z(\varphi, \theta)]$ (see \cref{sec:surface_description}). 

% This preference for low aspect ratio devices may compete with the quality particle confinement as it may be more difficult to obtain a good approximation of quasisymmetry on large volumes.
% Designing a stellarator aims to find a surface shape $S$ so that the magnetic field has a number of possibly competing physics and engineering properties.  
% The goal of a stellarator is to confine fast ions in the plasma and one way to achieve this is to design one that has the quasisymmetry property.
% A device that has a good quasisymmetry and ion confinement on a magnetic surface has a field strength that only depends on a linear combination of the two angles that parametrize the surface (Figure \ref{fig:...} right).

Without special shaping of the magnetic field, the plasma will not stay confined to the magnetic surfaces -- plasma particles will drift off surfaces until they hit the chamber walls. To improve fusion performance, and prevent damage to plasma facing components, the magnetic field is optimized to confine the plasma particles. In addition to particle confinement, goals of stellarator design include turbulence and instability suppression, controlled heat load on the plasma facing components, and extraction of \say{ash} particles. 
Stellarator optimization problems can be stated as PDE-constrained optimization problem of the form,
\begin{equation}
\begin{aligned} \label{eq:minimize}
        \min_{\xb, \Bb} & ~J(\Bb, \xb), \\
        \text{subject to: } & \eqref{eq:idealMHD}, 
        \\
        &\cb(\Bb, \xb) = 0,
\end{aligned}
\end{equation}
where the objective, $J$, and constraints, $\cb$, encode desirable stellarator properties.
Practically, \Cref{eq:minimize} can be solved by eliminating $\Bb$ via the PDE constraint \eqref{eq:idealMHD}, and minimizing the reduced objective $\hat J(\xb) = J(\Bb(\xb), \xb)$ subject to the reduced constraints $\hat{\cb}(\xb) = \cb(\Bb(\xb), \xb) = 0$. 
% For gradient-based minimization algorithms, $\hat J, \nabla_{\xb}\hat J$ are evaluated by solving \eqref{eq:idealMHD} for $\Bb$ given a boundary shape $S(\xb)$.  The design constraints $\mathbf c$ in \eqref{eq:minimize} may be taken into account using a penalty method. 

Due to the complexity of stellarator design, it is common to split the problem into two stages: in stage-I the shape of the plasma boundary $S$ is optimized so that magnetic field has nested magnetic surfaces, good particle confinement, suppresses plasma instabilities, etc.; in stage-II the electromagnetic coils are designed to produce the magnetic field found in stage-I \cite{zhu_new_2017,wechsung_precise_2022}. % Both problems can be stated as a PDE-constrained optimization problem of the form,
% \begin{equation}
% \begin{aligned} \label{eq:minimize}
%         \min_{\xb} & ~J(\xb), \\
%         \text{subject to: } & \eqref{eq:idealMHD}, 
%         \\
%         &\cb(\xb) = 0.
% \end{aligned}
% \end{equation}
% In the stage-I problem, the decision variables, $\xb$, parameterize the shape of $S$, while in the stage-II problem the decision variables represent the currents and shapes of the electromagnetic coils.
% Like the stage-I problem is a physics problem, the stage-II problem is an engineering one.
The stage-II problem is a mechanical engineering problem, determined by constraints, $\cb$, on the shape of the coils (length, curvature, torsion) -- material properties and manufacturing practices place tolerances and limits on the shapes and currents which coils can sustain (see \cite{zhu_new_2017} for an introduction to stage-II). In certain settings, the stage-I and II are solved simultaneously as a \textit{single-stage} -- an end-to-end optimization \cite{giuliani_single-stage_2022,giuliani_direct_2022,giuliani_direct_2023,giuliani_direct_2024}.

In this work, we focus on solutions to the stage-I problem.
% The stage-I problem is a PDE-constrained optimization problem,
% \begin{equation}
% \begin{aligned} \label{eq:minimize}
%         \min_{\xb} & ~J(\xb), \\
%         \text{subject to: } & \eqref{eq:idealMHD}, 
%         \\
%         &\cb(\xb) = 0.
% \end{aligned}
% \end{equation}
The decision variables, $\xb$, parameterize the shape of the plasma boundary, $S$. 
Two common constraints, which we consider in this work, constrain the aspect ratio and $\meaniota$ of the device to target values $A^*, \meaniota^*$: $c_A(\xb) = \frac{A(\xb) - A^*}{A^*}$ and $c_\iota(\xb) = \frac{\meaniota(\xb) - \meaniota^*}{\meaniota^*}$ \cite{landreman_magnetic_2022, baillod_stellarator_2022,baillod_integrating_2025,jorge_single-stage_2023,landreman_mapping_2022}. The objective function used in this work, $J = J_{QS}$ \cite{giuliani_direct_2022}, measures the deviation of the magnetic field from the nearest quasisymmetric magnetic field, and is a proxy measure for particle confinement (see \Cref{sec:surface_description}). Quasisymmetry is a symmetry in the strength of the magnetic field, $\|\Bb\|$, that ensures that particle trajectories have no radial (outward) drift, thus ensuring good particle confinement \cite{helander_theory_2014,landreman_magnetic_2022}. Quasisymmetric magnetic fields can be identified by plotting the contours of the field strength in a special coordinate system, Boozer coordinates --  the field strength contours of quasisymmetric magnetic fields only depend  on a linear combination of the Boozer angles, $\|\Bb\|(\theta,\varphi) = \|\Bb\|(\theta - N\nfp\varphi)$, thus appearing straight in Boozer coordinates.
% When plotted in a special coordinate system, Boozer coordinates, the contours of the field strength quasisymmetric magnetic fields appear straight, i.e. $\|\Bb\|(\theta,\varphi) = \|\Bb\|(\theta - N\nfp\varphi)$ 
% One way to achieve good particle confinement is to design the surface $S$ such that the magnetic field is \textit{quasisymmetric}. 
For example, in \cref{fig:stellarator_example}(C) the fieldlines are approximately parallel to the $x$-axis (aside from the small curve near the center). The type of quasisymmetry is parameterized by the normalized \textit{helicity}, $N$: $N=0$ denotes quasiaxisymmetry (QA) in which $\|\Bb\|$ appears flat when plotted in Boozer coordinates e.g. \cref{fig:stellarator_example}(C); $N=1$ denotes quasi-helical symmetry (QH) in which the contours of $\|\Bb\|$ form diagonal lines when plotted in Boozer coordinates i.e. \cref{fig:contours_in_sample}(second row, column 3). Both types of quasisymmetry are desirable, however, there is an interplay between $\nfp$ and $N$ that makes it difficult to find high $\nfp$, QA devices, and low $\nfp$, QH devices. The metrics $c_\iota, c_A, J_{QS}$ can all be computed from the solution of \cref{eq:idealMHD}, see Appendix \ref{sec:surface_description} for details.

Challenges of the stage-I problem \cref{eq:minimize} include: high dimensionality ($\nx =\mathcal{O}(100-1,000)$); moderate computational cost ($5-10$ seconds to solve \cref{eq:idealMHD} or compute a gradient when using MPI-parallelism or GPU acceleration); multiple local minima.
% We can introduce symmetries to reduce the dimensionality of the problem, though the use of derivative-free algorithms still remains tricky for problem sizes of interest.Gradient-based optimization algorithms are crucial to find solutions of \eqref{eq:minimize}. If the ideal MHD solver does not have derivatives implemented, then finite differences have successfully been used.  DESC is a solver that relies on JAX to obtain derivatives automatically.
Due to the high dimensionality of $\xb$, gradient-based optimization algorithms are crucial to efficiently find solutions. Since \cref{eq:minimize} typically has multiple minima, globalization techniques can be useful for adequately exploring the landscape of solutions. Generating solutions comes at a high computational cost, and limits the extent of exploration.

Generative models are poised to overcome the computational challenges associated with sampling and optimization, since they can aggregate information across a variety of data sources and generate valuable solutions, rapidly. Generative models can also produce samples at new sets of conditions and provide a means of exploration of the stellarator landscape. To solve \cref{eq:minimize}, a generative model would learn a sampling mechanism to produce $\xb$ given conditions $\yb$. In this work, we characterize a (quasisymmetric) stellarator with the four conditions $\yb = (\meaniota, A, \nfp, N)$. A high quality generative model would consistently produce designs that minimize $J_{QS}(\xb)$ and satisfy $c_\iota(\xb) = 0, c_A(\xb) = 0$. Therefore, we use these metrics to evaluate the stellarators produced by a generative model.

\subsection{Related Work}
\label{sec:related_work}

\textbf{Fusion/stellarator optimization}
To our knowledge, the first work designing stellarators with generative models was \cite{curvo_using_2025}, which used a mixture density network \cite{bishop_pattern_2006} to generate stellarators with quasisymmetry in their core. Yet challenges remain to generate stellarators with quasisymmetry throughout their volume, and to improve adherence to conditions. \cite{cadena_constellaration_2025} trained a classifier to determine whether stellarator configurations satisfy a set of constraints and used Markov chain Monte Carlo methods to sample from this region. Another work, \cite{candido_design_2023} took a regression approach, learning a nonlinear regressor from conditions to features. \Cref{sec:related_work_extended} covers additional work using ML and generative models in fusion research.

\textbf{Generative models} 
To our knowledge, this is the first use of diffusion models \cite{ho_denoising_2020, nichol_improved_2021} in stellarator design. Diffusion models have been applied successfully for solving inverse problems \cite{chung_diffusion_2022,chung_improving_2022,daras_survey_2024}, and widely used for generative modeling in astrophysics, medical imaging \cite{mudur_can_2022, remy_generative_2023}, image generation \cite{rombach_high-resolution_2022}, audio synthesis \cite{kong_diffwave_2020}, and protein design \cite{watson2023novo}. Generative adversarial networks and variational autoencoders are also a common choice for solving inverse problems \cite{asim_invertible_2020,zhao_generative_2023, goh_solving_2019}, and generating solutions to problems in material science and molecular design \cite{choudhary_recent_2022,gomez-bombarelli_automatic_2018}. 
% Models with custom architecture have also been used \cite{jumper2021highly}, material design \cite{merchant2023scaling}. 
% Diffusion models have demonstrated impressive generation quality and favorable optimization performance compared to other generative models such as generative adversarial networks (GAN). 

%
\section{Case Study: Designing Stellarators With Diffusion Models}
% \section{A case study in designing stellarators with diffusion models}
\label{sec:case_study}
In this section, we share the methods and results of a case study in which diffusion models are trained to solve \nameref{prob:main}. Specifically, our aim is to train a diffusion model to generate boundary shapes, $S$, of \textit{quasisymmetric} stellarators. Generating boundary shapes for quasisymmetric stellarators would solve the stage-I stellarator optimization problem. 
Our code and data are available on Zenodo \cite{padidar_diffusion_2025}. 

\begin{figure}[tbh!]
    \centering
    \includegraphics[width=1.1\textwidth]{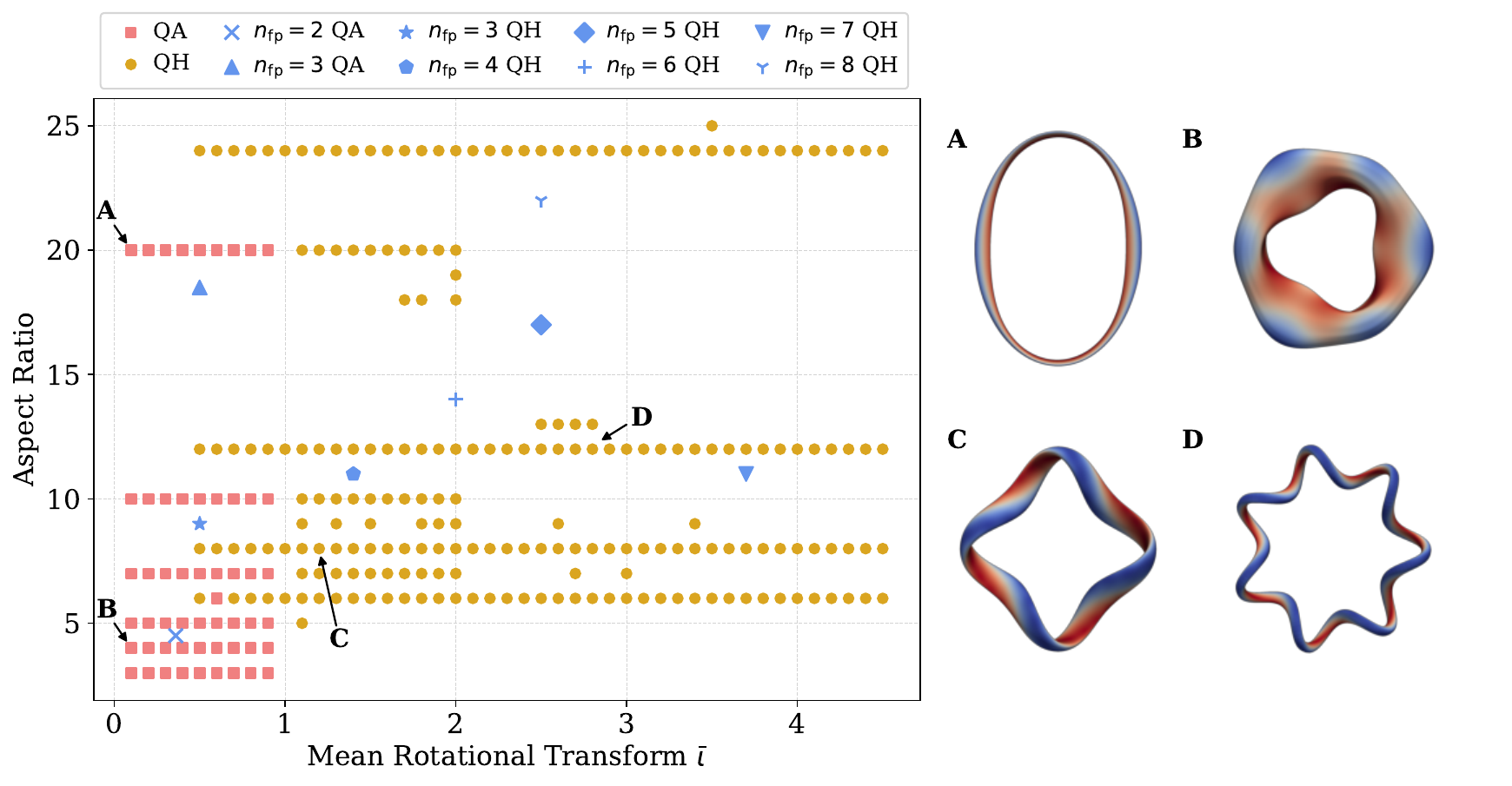}
    \caption{(Left) Values of the mean rotational transform and aspect ratio for devices in QUASR, colored by the type of quasisymmetry: QA (red square), QH (gold circle). The various blue markers indicate arbitrary conditions that are not within QUASR; in \Cref{fig:contours_in_sample,fig:out_of_sample_boxplot}, the diffusion model is used to generate stellarators with these conditions. (Right) Visualization of four devices in QUASR: (A) an $\nfp=2$-QA device; (B) an $\nfp=3$-QA device; (C) an $\nfp=4$-QH device; (D) an $\nfp=7$-QH device. The mean rotational transform and aspect ratio of each configuration can be seen on the left figure (labels A, B, C, D). 
    }
    \label{fig:quasr_data_range}
\end{figure}

\textbf{Training data}
% \label{sec:quasr_data}
%
We use data from the \href{https://quasr.flatironinstitute.org/}{QUAsisymmetric Stellaratator Repository (QUASR)}. With over 300,000 stellarators, the QUASR repository is a large database of optimized stellarators including both coils and magnetic surfacs. QUASR was generated using Bayesian global optimization over the space of vacuum (current-free) stellarators, in search of quasisymmetric stellarators with a range of $n_{\text{fp}}$, aspect ratio $A$, and rotational transform $\iota$ \cite{giuliani_comprehensive_2025, giuliani_direct_2024}. 

While the devices in QUASR often have multiple associated magnetic surfaces, we construct our dataset by just taking the outermost surface. The feature vector, $\xb$ consists of $\nx=661$ Fourier coefficients describing the Cartesian components of the surface $S$, $[x(\varphi, \theta), y(\varphi, \theta), z(\varphi, \theta)]$, through three separate two-dimensional Fourier series in the poloidal and toroidal Boozer angles (see \Cref{sec:surface_description} for more details). Fourier features have a natural scaling to them: features corresponding to high mode number Fourier coefficients have less impact on the surface geometry than those with low mode number. Since the feature representation of each surface is not unique, the surfaces are \say{canonicalized} to a unique form, using the procedure in \cite{giuliani_comprehensive_2025}. Since the initial submission of this manuscript, it was found that less than 1\% of the data points were affected by an error in the canonicalization procedure and roughly 8\% of the data appeared twice in the dataset. Retraining the model on a corrected dataset showed that the main results were unaffected by these errors. For transparency, both datasets have been made available.

The diffusion model is given $\ny=4$ conditions, $\yb = (\meaniota, A, N, \nfp)$. $J_{QS}$ is not treated as a condition since the stage-I problem seeks devices with $J_{QS}=0$ and specifying a value of $J_{QS}$ as a condition is unintuitive for domain specialists. The distribution of $\meaniota$ and aspect ratio of the devices in QUASR is shown in \Cref{fig:quasr_data_range} as gold circles and red squares; the gold circles denote QH devices ($N=1$), while the red square denotes QA devices ($N=0$). 
% Four devices from the dataset are shown in the right panel. 
Notably, the devices sparsely sample the space of conditions, providing plenty of opportunity for the diffusion model to generate novel designs. The 8 blue markers in \Cref{fig:quasr_data_range} highlight arbitrary conditions that are \textit{not} in QUASR, which we use to evaluate the performance of the diffusion model in \Cref{fig:out_of_sample_boxplot}.

\textbf{Evaluation criteria}
A well-trained diffusion model should produce samples $\xb$ given conditions $\yb$, such that $c_\iota(\xb) = 0, c_A(\xb) = 0$. In addition, since the stellarators in the training data have low $J_{QS}$, we expect a well trained diffusion model to generate stellarators with similarly low values of $J_{QS}$. We quantify the performance of the diffusion model with three metrics: $c_\iota, c_A, J_{QS}$. Since mean-rotational transform and aspect ratio are conditions, $c_\iota, c_A$ measure the relative error in a $\meaniota(\xb), A(\xb)$ from the condition values.  We consider $c_{\iota}, c_{A} < 5\%$ to be satisfactory values of the metric. $J_{QS}$ on the other hand, measures the quasisymmetry error of $\xb$ rather than error from a condition. Quasisymmetry must be achieved to a relatively high tolerance to achieve good particle confinement \cite{wiedman_coil_2024}, so we consider $J_{QS} < 1\%$ to be a satisfactory value. See \Cref{sec:surface_description} for details on our evaluation procedure.

\textbf{Diffusion model}
%
% Our aim is to use a diffusion model to generate stellarator design variables, $\xb$, from an array of desirable characteristics, $\yb$: instead of learning $p(\xb)$, we want to learn $p(\xb|\yb)$. 
% The \textit{conditional} diffusion model learns a conditional density $p(\xb|\yb)$. 
% We follow the approach used in \cite{ho2020denoising} to modify DDPM to enable \say{conditional diffusion}. 
% The algorithm is an instance of DDPM \cite{ho2020denoising}, modified to to enable \say{conditional diffusion}. 
The \textit{conditional} diffusion model is an instance of a Denoising Diffusion Probabilistic Model (DDPM) \cite{ho_denoising_2020}, modified to learn a conditional density $p(\xb | \yb)$.
Following \cite{ho_denoising_2020}, we modify the neural network $f_\theta$ to accept the additional conditions $\yb$, i.e. $f_\theta(\xb_t,t,\yb)$. 
A detailed discussion of the training and sampling algorithms can be found in \Cref{sec:model_training}.

% %
% \begin{minipage}[t]{0.45\textwidth}
% \begin{algorithm}[H]
% \footnotesize
% \caption{Training $f_{\theta}$}
% \label{alg:ddpm_training}
% \KwIn{Step size sequence $\{\alpha_t\}_{t=1}^T$}
% % \KwOut{$N\times d$ matrix of samples $\hat{\Xb}$}
% \Repeat{converged}{
%  $\xb_0 \sim q\left(\xb_0\right), \zb \sim \mathcal{N}(0, I)$ \;
%  $t \sim \operatorname{Uniform}(\{1, \ldots, T\})$  \;
% Take a gradient descent step using
% \scriptsize{$\nabla_\theta\left\|\zb-f_\theta\left(\sqrt{\bar{\alpha}_t} \xb_0 + \sqrt{1-\bar{\alpha}_t} \zb , t, \yb\right)\right\|^2$}}
% \end{algorithm}
% %
% \end{minipage}
% \hspace{1em}
% \begin{minipage}[t]{0.5\textwidth}
% \begin{algorithm}[H]
% \footnotesize
% \caption{Sampling from trained $f_{\theta}$}
% \label{alg:ddpm_sampling}
% \KwIn{$f_\theta$, step sizes $\{\alpha_t\}_{t=1}^T$, variances $\{\sigma_t\}_{t=1}^T$}
% $\xb_T \sim \mathcal{N}(0, I)$ \;
% \For{$t = T, \ldots, 1$}{
% \scriptsize{$\zb \sim \mathcal{N}(0,I)$ if $t>1$ else $\zb=0$ \;}
% \scriptsize{$\xb_{t-1}=\frac{1}{\sqrt{\alpha_t}}(\xb_t-\frac{1-\alpha_t}{\sqrt{1-\bar{\alpha}_t}} f_\theta\left(\xb_t, t, \yb\right))+\sigma_t \zb$}
% }
% \Return $\xb_0$
% \end{algorithm}
% \end{minipage}
% %

\begin{figure}[tbh!]
    \centering
    \includegraphics[width=\textwidth]{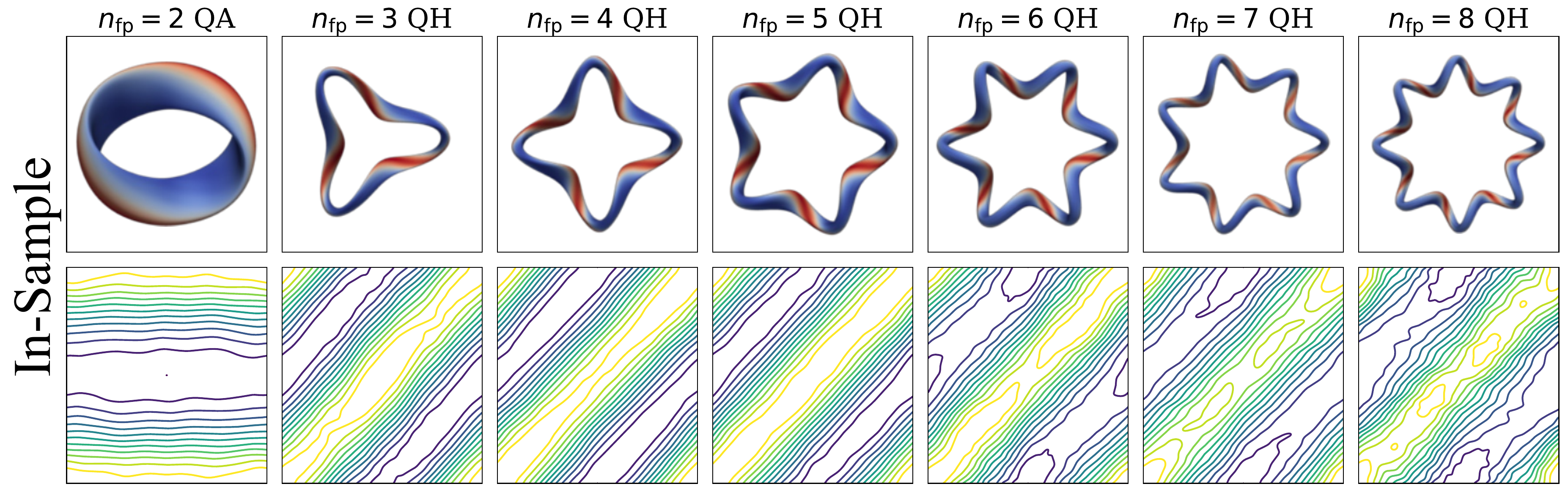}
    \includegraphics[width=\textwidth]{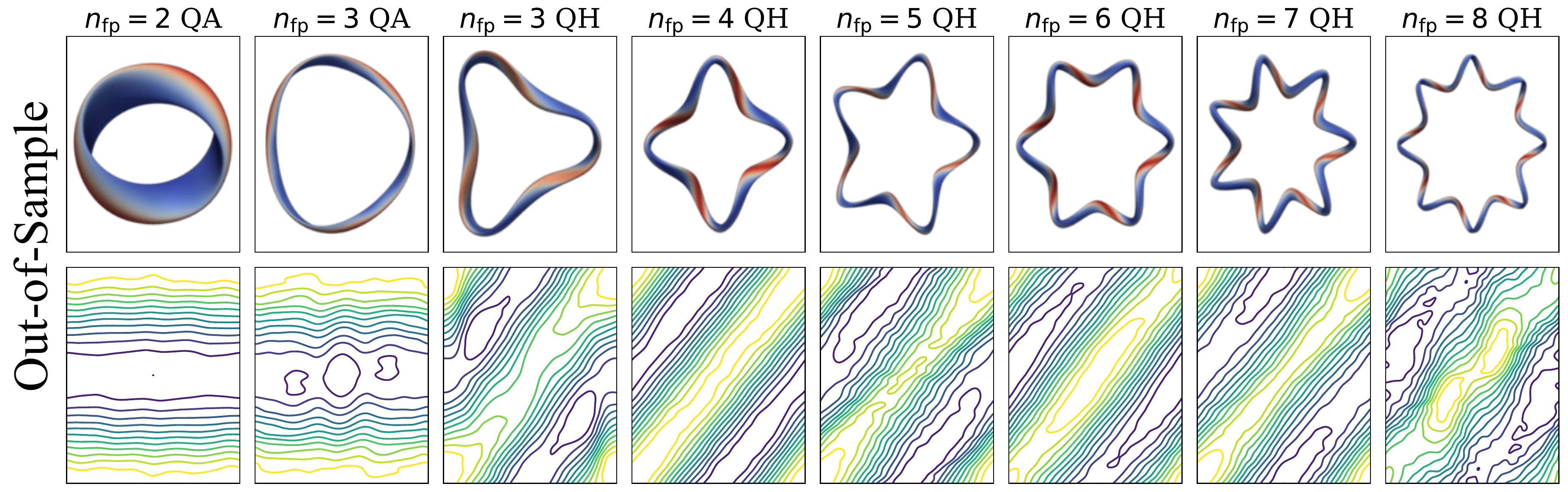}
    \caption{Stellarators generated by the diffusion model using in-sample conditions (first row) and out-of-sample conditions (third row). The contours of the magnetic field strength in Boozer coordinates of the in-sample devices (second row) and out-of-sample devices (fourth row), demonstrating the quality of quasisymmetry (perfectly straight contours would indicate perfect quasisymmetry) -- all in-sample (out-of-sample) devices have less than $2.5\%$ ($6\%$) deviation from quasisymmetry, and $5\%$ constraint violation, $c_\iota, c_A$. Conditions used to generate the devices are shown in \Cref{table:out_of_sample_conditions}. Each out-of-sample device shown here corresponds to a blue marker on \Cref{fig:quasr_data_range} with identical label.}
    \label{fig:contours_in_sample}
\end{figure}

%
% \subsection{In-sample performance}
\subsection{Generation Results}
In this section, we examine the performance of the diffusion model by evaluating the metrics $c_\iota, c_A, J_{QS}$ on samples $\xb$ drawn from the diffusion model. We consider two settings: the \say{in-sample} setting where conditions $\yb$ are taken directly from the training data, and the \say{out-of-sample} setting where the conditions $\yb$ do \textit{not} exist in the training data. Devices sampled with out-of-sample conditions are fundamentally new stellarator designs.

\Cref{fig:contours_in_sample} (top two rows) shows seven devices, each with a different number of field periods, generated by the diffusion model using conditions taken directly from the training data, while \Cref{fig:contours_in_sample} (bottom two rows) shows eight devices generated by the diffusion model using out-of-sample conditions. 
% The conditions used to generateare shown in \Cref{table:out_of_sample_conditions}. 
\Cref{fig:contours_in_sample} (rows 1 and 3) show top-down images of the generated stellarators, where the surfaces are colored by the field strength. \Cref{fig:contours_in_sample} (rows 2 and 4) show the contours of the magnetic field strength in Boozer coordinates on the surface. The near straight contours indicate good quasisymmetry in all devices: all in-sample devices have less than $2.5\%$ deviation from quasisymmetry, and out-of-sample devices have less than $6\%$ deviation from quasisymmetry. The good out-of-sample performance indicates the model learned the symmetry from the training data, and was able to extrapolate it to out-of-sample devices. \Cref{fig:contours_in_sample} illustrates that the diffusion model can generate sensible stellarators.

\begin{figure}[tbh!]
    \centering
    \includegraphics[width=1.0\textwidth]{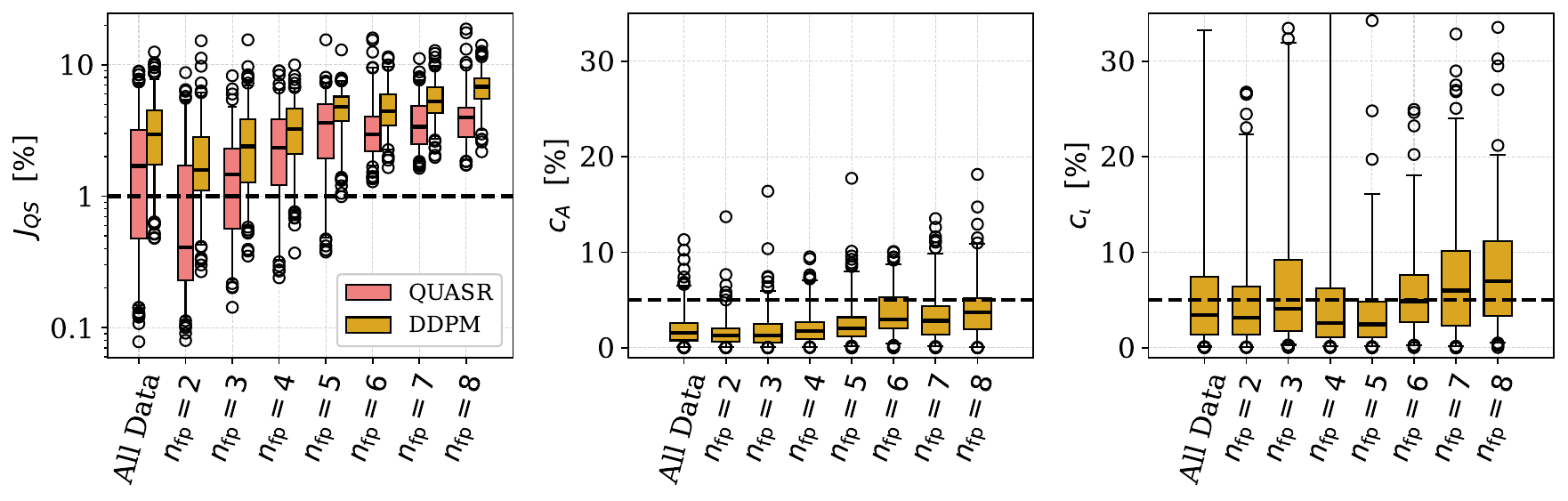}
    \caption{Performance of stellarators sampled by the diffusion model given in-sample conditions. The labels $\nfp=k$ shows the performance of the subset samples with the condition $\nfp=k$. (Left) Quasisymmetry error of samples from DDPM compared to the quasisymmetry of stellarators in QUASR (labeled Actual). DDPM generates stellarators with levels of $J_{QS}$ similar to that of QUASR, even though quasisymmetry was not used as a condition to train the diffusion model. (Middle/Right) Percentage error in the aspect ratio (middle) and mean rotational transform (right) from the aspect ratio (mean rotational transform) condition used to generate the sample. In all panes, the black dashed line denotes a desirable value of the metric: $J_{QS} \leq 1\%$ and $c_\iota, c_A \leq 5\%$.}
    \label{fig:boxplot_in_sample}
\end{figure}

\Cref{fig:boxplot_in_sample} precisely quantifies the performance of the diffusion model in generating stellarators from \emph{in-sample} conditions. Each frame shows box plots for a different metric: $J_{QS}$ (left), $c_A$ (middle), $c_\iota$ (right). In each frame, the tick labels on the $x$-axis are used to indicate different subsets of the data, e.g. the boxes above $\nfp=2$ show the performance of DDPM when only given conditions with $\nfp=2$. The left frame contrasts the distribution of $J_{QS}$ for devices generated by the diffusion model (gold boxes labeled DDPM) to that of devices from the training data (red boxes labeled QUASR). The devices generated by DDPM have similar, though somewhat worse quasisymmetry than the devices in QUASR, but are still near the target value of $1\%$ (black dashed line). The middle and right frames show $c_A, c_\iota$ for devices drawn from DDPM. For all subsets of the data, the aspect ratio of generated devices is typically less than $5\%$ from the aspect ratio condition. On the other hand, $\meaniota$ tends to have higher error from the condition, coming within $12\%$ of the condition, $75\%$ of the time, for all subsets of the data. Overall, the in-sample performance of the diffusion model is satisfactory, hitting the target error values the majority of the time, while leaving room for improvement in future work.

\begin{figure}[tbh!]
    \centering
    \includegraphics[width=\textwidth]{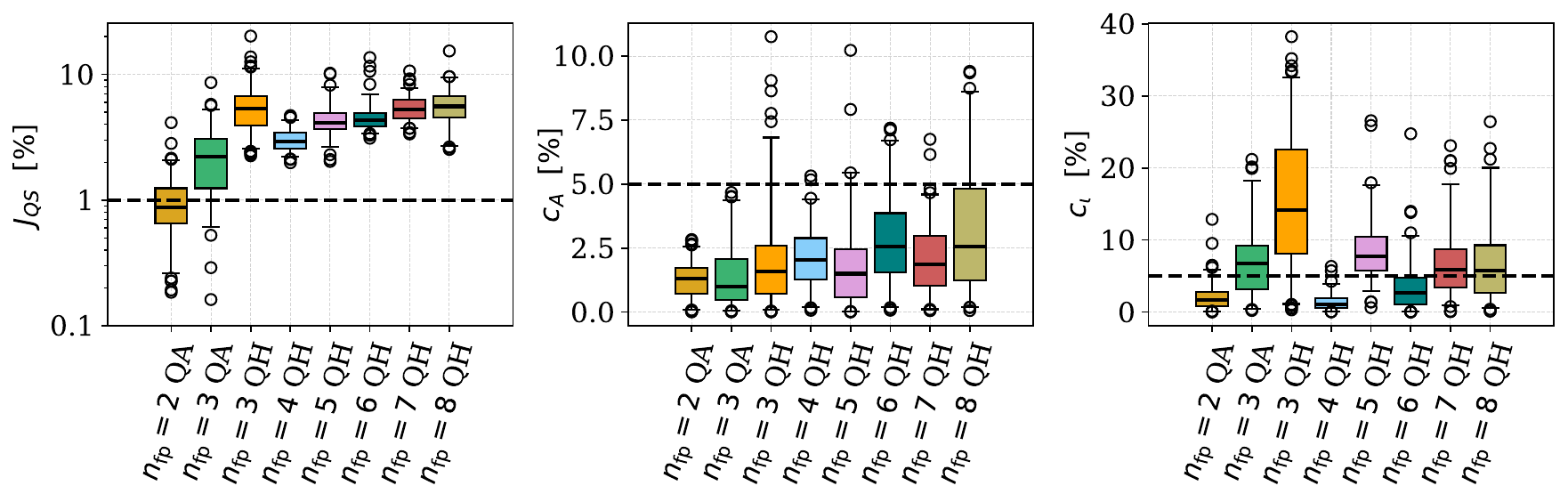}
    \caption{
    Performance of stellarators generated by the diffusion model given \textit{out-of-sample} conditions, in terms of $J_{QS}$ (left), $ c_A$ (middle), $c_\iota$ (right). $J_{QS}$ measures quasisymmetry deviation of devices, whereas $c_A, c_\iota$ measure the relative error of $A, \meaniota$ from the condition used to generate the device. Black dashed lines indicate desirable value of the metric: $J_{QS} \leq 1\%$ and $c_\iota, c_A \leq 5\%$. Each box corresponds to a single condition from \Cref{table:out_of_sample_conditions}. The conditions are also plotted as a blue marker on \Cref{fig:quasr_data_range} with identical label.}
    \label{fig:out_of_sample_boxplot}
\end{figure}

\Cref{fig:out_of_sample_boxplot} precisely quantifies the performance of the diffusion model in generating stellarators, with the \emph{out-of-sample} conditions from \Cref{table:out_of_sample_conditions}. Each figure shows box plots for a different metric: $J_{QS}$ (left), $c_A$ (middle), $c_\iota$ (right). In each frame, the tick labels on the $x$-axis are used to indicate which of the eight out-of-sample conditions was used, e.g. the boxes above $(\nfp=2 \text{ QA})$ show the performance of DDPM when only given $(\nfp=2, N=0)$ out-of-sample condition from \Cref{table:out_of_sample_conditions}. The left panel shows that the deviation from quasisymmetry for each of the eight conditions is less than $10\%$, and in some cases less than the $1\%$ target. The middle panel shows that for all conditions, the aspect ratio of generated devices is seldom more than $5\%$ from the aspect ratio condition. Similarly the right panel shows that the $\meaniota$ of generated devices tends to be similar to the condition, coming within $10\%$ of the condition, $75\%$ of the time, for all conditions except $\nfp=3$, QH. While there may be room to improve the precision of the diffusion model, \Cref{fig:out_of_sample_boxplot,fig:contours_in_sample} concretely demonstrate that quasisymmetric devices can indeed be generated out-of-sample. 

\section{Open Research Problems}
\label{sec:open_problems}
We now discuss variants and extensions of \nameref{prob:main} that are open areas for research. 

\textbf{Dataset and Benchmark Curation} Despite the availability of large-scale stellarator datasets, no standard benchmark exists to unify them and provide consistent evaluation protocols for method comparison. A promising direction is to curate such benchmark by considering the following diverse stellarator datasets across stellarator classes, data representations, and properties: %; each covers distinct classess of stellarators, uses potentially different representations, and contains different sets of properties (conditions). 
the QUASR database [\href{https://quasr.flatironinstitute.org/}{link}] contains coils and magnetic surfaces of quasisymmetric stellarators; the Constellaration dataset \cite{cadena_constellaration_2025} contains surfaces for quasi-isodynamic stellarators; the databases generated in \cite{landreman_mapping_2022,curvo_using_2025} contain the near-axis representation of quasisymmetric stellarators; the omnigenity database \cite{gaur_omnigenous_2024} contains surfaces for omnigenous stellarators; and the database from \cite{landreman_how_2025} contains surfaces for stellarators with measures of turbulence.

\textbf{Physics-informed Architecture, Optimization, and Sampling} In our case study (\cref{sec:case_study}), the diffusion model learned to generate approximately quasisymmetric stellarators purely from the training data. A promising direction is to incorporate physical constraints explicitly in the design of the generative model architecture (e.g. equivariant models), the loss function, or the sampling procedure. For example, replacing the non-differentiable evaluator \texttt{VMEC}, with a differentiable evaluator such as \texttt{DESC} \cite{dudt_desc_2020} would enable the use of \say{physics informed loss functions} which penalize deviation from desired physical properties, i.e. penalize $J_{QS}, c_A, c_\iota$ \cite{bastek_physics-informed_2025}.
Similarly, derivatives of physics properties could be used in guided sampling procedures to ensure
generated devices have, for instance, low $J_{QS}, c_A, c_\iota$ \cite{dhariwal_diffusion_2021}. Derivative information is also commonly available in the \say{single-stage} stellarator design setting \cite{giuliani_direct_2022,giuliani_single-stage_2022}, where a generative model could be trained to generate a coil set-magnetic surface pair. Generation quality can also be improved by combining multiple data sources into a multi-fidelity generative model. For example, a multi-fidelity model for turbulent heat flux could combine low-fidelity analytic models,
% \cite{landreman2025does, roberg2023critical,proll2015tem,mackenbach2022available,mynick2010optimizing,mynick2014turbulent,hegna2018theory,nakata2022impact,nakayama2023simplified,goodman2024quasi},
with sparse, high-fidelity data \cite{landreman_how_2025}.

\section{Acknowledgments}
The author would like to thank the Flatiron Institute’s Scientific Computing Core for their support, in addition to Jiequn Han for his helpful feedback.

\bibliography{references/mp_ref_zotero,references/teresa_ref}
\bibliographystyle{abbrv}

%%%%%%%%%%%%%%%%%%%%%%%%%%%%%%%%%%%%%%%%%%%%%%%%%%%%%%%%%%%%
\clearpage
\appendix

\section{Description and Evaluation of Stellarator Designs}
\label{sec:surface_description}
In this section, we discuss the representation of surfaces and evaluation of $J_{QS}, c_A, c_\iota$. Surfaces can be constructed with \texttt{SIMSOPT} \cite{landreman_simsopt_2021} and evaluated with \texttt{VMEC}. Some stellarators in the training data, or generated by the diffusion model, have highly shaped boundaries. \texttt{VMEC} struggles to evaluate these, failing to return a solution for approximately $41\%$ of stellarators from the training data, and $57\%$ of stellarators generated in-sample by the diffusion model. In reporting results, we ignore these designs.

\paragraph{Surface Parameterization}
We follow \cite{giuliani_direct_2022,giuliani_direct_2024} in describing surfaces with the \texttt{SurfaceXYZTensorFourier} format in \texttt{SIMSOPT}. The vector $\xb$ is the set of Fourier coefficients $x_{ij}, y_{ij}, z_{ij}$ used in \Cref{eq:surface_x,eq:surface_y,eq:surface_z}, with ordering dictated by \texttt{SIMSOPT}. Stellarator symmetry is assumed for all surfaces in training and evaluation.
The Cartesian components of a $\nfp$-field period surface, $x(\varphi,\theta), y(\varphi, \theta), z(\varphi,\theta)$, can be constructed with the Fourier series,
\begin{align}
    \hat x(\varphi, \theta) &= \sum_{i=0}^{m_\text{pol}} \sum_{j=0}^{n_\text{tor}} x_{ij} w_i(\theta)v_j(\varphi) + \sum_{i=m_\text{pol}+1}^{2m_\text{pol}} \sum_{j=n_\text{tor}+1}^{2n_\text{tor}} x_{ij} w_i(\theta)v_j(\varphi)
    \\
     \hat y(\varphi, \theta) &= \sum_{i=0}^{m_\text{pol}} \sum_{j=n_\text{tor}+1}^{2n_\text{tor}} y_{ij} w_i(\theta)v_j(\varphi) + \sum_{i=m_\text{pol}+1}^{2m_\text{pol}} \sum_{j=0}^{n_\text{tor}} y_{ij} w_i(\theta)v_j(\varphi)
     \\
    x(\varphi, \theta) &= \hat x(\varphi, \theta)  \cos(\varphi) - \hat y(\varphi, \theta)  \sin(\varphi)
    \label{eq:surface_x}
    \\
    y(\varphi, \theta) &= \hat x(\varphi, \theta)  \sin(\varphi) + \hat y(\varphi, \theta)  \cos(\varphi)
    \label{eq:surface_y}
    \\
    z(\varphi, \theta) &= \sum_{i=0}^{m_\text{pol}} \sum_{j=n_\text{tor}+1}^{2n_\text{tor}} z_{ij} w_i(\theta)v_j(\varphi) + \sum_{i=m_\text{pol}+1}^{2m_\text{pol}} \sum_{j=0}^{n_\text{tor}} z_{ij} w_i(\theta)v_j(\varphi)
    \label{eq:surface_z}
\end{align}
where the basis functions $v_j$ are,
\begin{align}
    \{1, \cos(1\,\mathrm{nfp}\,\varphi), \ldots, \cos(n_\text{tor}\,\mathrm{nfp}\,\varphi), \sin(1\,\mathrm{nfp}\,\varphi), \ldots, \sin(n_\text{tor}\,\mathrm{nfp}\,\varphi)\},
\end{align}
and the basis functions $w_i$ are,
\begin{equation}
    \{1, \cos(1\theta), \ldots, \cos(m_\text{pol}\theta), \sin(1\theta), \ldots, \sin(m_\text{pol}\theta)\}.
\end{equation}
The training data dictated that $m_{\text{pol}}=n_{\text{tor}}=10$.

\paragraph{Quasisymmetry Objective}
The quasisymmetry objective, $J_{QS}$, can be evaluated from a magnetic field, $\Bb$, and surface parameterized in Boozer coordinates $\theta, \varphi$. 
The $J_{QS}$ objective from \cite{giuliani_direct_2022} computes the relative distance between the quasisymmetric and non-quasisymmetric components of the field strength $B(\varphi,\theta) =  \| \mathbf B(\varphi,\theta) \|_2$,
\begin{equation}
    J_{QS} = \sqrt{\frac{\int_{\Gamma_{s}} B_{\text{non-QS}}^2~dS}{\int_{\Gamma_{s}} B_{\text{QS}}^2~dS}}.
\end{equation}
$B_{\text{QS}}$ is the projection of $B$ onto the space of quasisymmetric fields. For quasi-axisymmetry,
\begin{align}
    B_{\text{QS}} &= \frac{\int_0^1 B \|\mathbf n\| ~d\varphi}{\int_0^1 \|\mathbf n\| ~d\varphi} \\
    B_{\text{non-QS}} &= B - B_{\text{QS}}
\end{align}
For quasihelical symmetry, the integral is taken over the helical angle $\theta - N\nfp\varphi$. When $J_{QS}$ is zero, the magnetic field is perfectly quasisymmetric on the magnetic surface.

\paragraph{Evaluation}
The metrics $J_{QS}, c_\iota, c_A$ are evaluated with the following protocol. Given $\nfp$ and $\xb$ a surface is object is constructed in \texttt{SIMSOPT}. The surface is evaluated with \texttt{VMEC}, from which $\meaniota$ and $A$ can be computed. Boozer coordinates are constructed using the \texttt{BoozXform} in \texttt{SIMSOPT}. $B$ and the jacobian $\sqrt{g}$ are constructed in Boozer coordinates, from which $J_{QS}$ can be evaluated.

\section{Diffusion Model Details}
\label{sec:model_training}
The \textit{conditional} diffusion model is an instance of the DDPM model \cite{ho_denoising_2020} where the neural network model, $f_\theta$, is modified to enable conditions. Training and sampling followed the standard DDPM training and sampling algorithms, \Cref{alg:ddpm_training,alg:ddpm_sampling}. \Cref{alg:ddpm_training} requires as input a sequence of step sizes $\{\alpha_t\}_{t=0}^T$ and a neural network model $f_\theta(\xb_t,t,\yb)$ which predicts a noisy sequence given a (noised) feature $\xb_t$, timestep $t$, and condition $\yb$. The conditions and noise-free features $\yb\sim q(\yb), \xb_0\sim q(\xb_0 ~|~\yb)$ are drawn from the training data.

The model was trained with $\ny=4$ features, $\yb = (\meaniota, A, \nfp, N)$. The diffusion model used $200$ timesteps, with a linear schedule. A standard feed-forward neural network architecture was used to describe $f_\theta(\xb_t, t, \yb)$ with $4$ hidden layers, each with width $2048$, and Gaussian error linear unit (GELU) activiation functions. A sinusiodal network head was used to embed $\xb,t, \yb$ before entering the standard feed-forward layers. The network head maps $\xb$ to dimension $64$, $t$ to dimension $128$, and $\yb$ to dimension $128$. The \texttt{Adam} optimizer \cite{kingma_adam_2017} was used with a training batch size of $4096$, an evaluation batch size of $128$, for $250$ epochs, with a learning rate of $0.0005$.

\begin{minipage}[t]{0.45\textwidth}
\begin{algorithm}[H]
\small
\caption{Training $f_{\theta}$}
\label{alg:ddpm_training}
\KwIn{Step size sequence $\{\alpha_t\}_{t=1}^T$}
% \KwOut{$N\times d$ matrix of samples $\hat{\Xb}$}
\Repeat{converged}{
 $\yb  \sim q\left(\yb\right)$, $\xb_0 \sim q\left(\xb_0 | \yb\right), \zb \sim \mathcal{N}(0, I)$\;
 $t \sim \operatorname{Uniform}(\{1, \ldots, T\})$  \;
Take a gradient descent step using
$\nabla_\theta\left\|\zb-f_\theta\left(\sqrt{\bar{\alpha}_t} \xb_0 + \sqrt{1-\bar{\alpha}_t} \zb , t, \yb\right)\right\|^2$}
\end{algorithm}
\end{minipage}
\hspace{1em}
\begin{minipage}[t]{0.5\textwidth}
\begin{algorithm}[H]
\small
\caption{Sampling from trained $f_{\theta}$}
\label{alg:ddpm_sampling}
\KwIn{$f_\theta$, step sizes $\{\alpha_t\}_{t=1}^T$, variances $\{\sigma_t\}_{t=1}^T$, conditions $\yb$.}
$\xb_T \sim \mathcal{N}(0, I)$ \;
\For{$t = T, \ldots, 1$}{
$\zb \sim \mathcal{N}(0,I)$ if $t>1$ else $\zb=0$ \;
$\xb_{t-1}=\frac{1}{\sqrt{\alpha_t}}(\xb_t-\frac{1-\alpha_t}{\sqrt{1-\bar{\alpha}_t}} f_\theta\left(\xb_t, t, \yb\right))+\sigma_t \zb$
}
\Return $\xb_0$
\end{algorithm}
\end{minipage}

\paragraph{Dimensionality reduction} 
\label{sec:dimensionality_reduction}

\begin{figure}[tbh!]
    \centering
    \includegraphics[width=\textwidth]{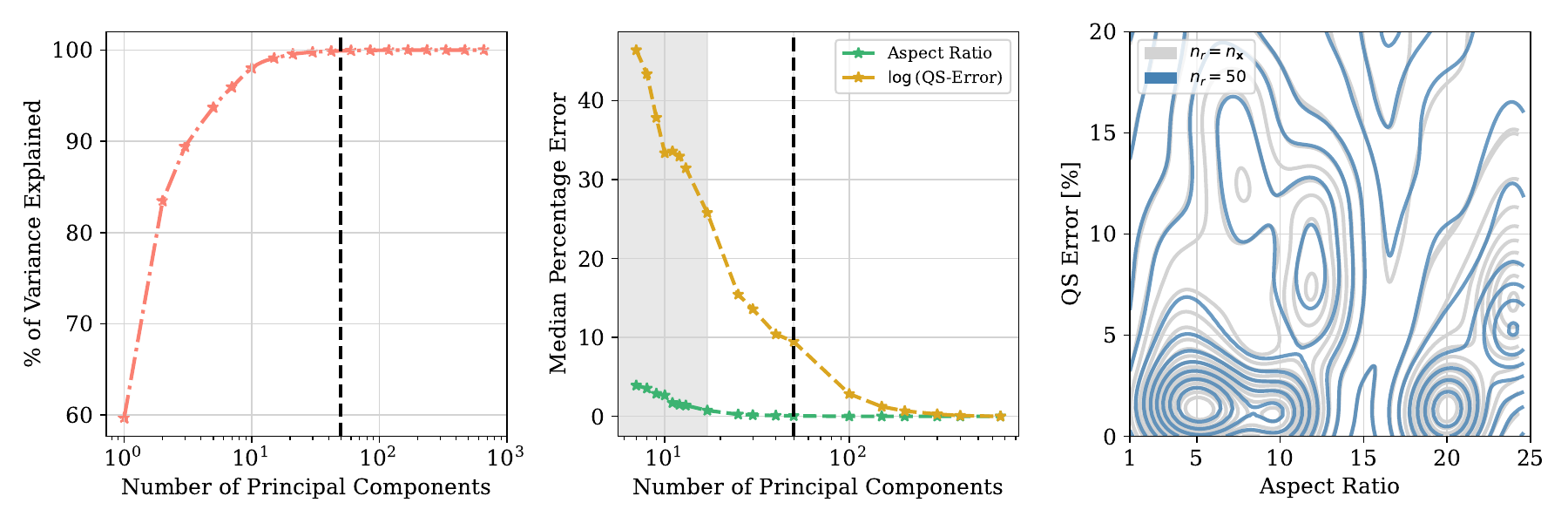}
    \caption{(Left) The fraction of the variance of $\xb$ explained by PCA with a given embedding dimension, $n_r$. The black dashed line marks an embedding dimension of $50$. (Middle) The median percentage error of $\log(J_{QS})$ and aspect ratio between embedded devices and devices in the dataset. The properties of devices can change significantly as the dimension of the embedding is reduced -- the grey region highlights areas where the \textit{max} relative error in the aspect ratio is at least 20\%. (Right) Joint probability density of the quasi-symmetry error and aspect ratio of devices embedded in $n_r=50$ dimensions (blue) versus devices in the actual dataset (grey). Using $n_r =50$ ensures that significant information is \textit{not} lost by dimensionality reduction.}
    \label{fig:dimensionality_reduction}
\end{figure}

Prior to training the diffusion model, we reduce the dimensionality of the raw training data to eliminate noise and high frequency oscillations in the dataset that do not have a meaningful connection to stellarator performance. Using principal component analysis (PCA), we project the data points, $\xb$, onto the linear subspace of dimension $\nr <\nx$ which captures largest fraction of the data variance. As the dimension of the embedding reduces, so does the quality of the devices in the embedded dataset. Missing important information in the training set, the diffusion model is limited in what it can learn, and the quality of stellarators it will generate. \Cref{fig:dimensionality_reduction} shows the reduction in the quality of stellarators in the dataset as the embedding dimension decreases. There is not a significant loss in performance of devices for the embedding dimension used for training, $n_r = 50$. Prior to evaluating a sample from the diffusion model, the sample must be projected back up to the full $\nx$-dimensional space.

\section{Additional Related Work}
\label{sec:related_work_extended}
Besides the work on \nameref{prob:main}, there have been many approaches to use ML to understand fusion plasmas. 
\cite{landreman_how_2025,sengupta_periodic_2025} used symbolic regression to understand turbulence and quasisymmetry in stellarators. \cite{laia_data-driven_2025} explored correlations in a dataset of stellarators and developed predictive models for quasisymmetry. 
\cite{liu_diff-pic_2024} used diffusion models to accelerate particle-in-cell simulations. 
\cite{clavier_generative-machine-learning_2025} used a variational autoencoder (VAE) to generate time-varying solutions to the Hasagawa-Wakatani equations for turbulence. \cite{nieuwenhuizen_modeling_2021} used VAEs to estimate hidden parameters that influence tokamak operation. \cite{wan_high-fidelity_2025} pre-trained a model for tokamak turbulence on low-fidelity data and fine-tune on high-fidelity data, developing an multi-fidelity surrogate for turbulence simulations. \cite{van_rijn_minimizing_2022,vos_discovery_2024} trained a VAE as a surrogate for particle and heat flux. \cite{wakasa_construction_2007} constructed a neural network surrogate for neoclassical transport coefficients in the LHD device. Physics-regularized neural networks have been used to predict solutions to the ideal MHD equations in W7-X geometries \cite{merlo_physics-regularized_2023,merlo_proof_2021}. \cite{pavone_machine_2023} reviewed applications of Bayesian and ML methods in fusion research, spanning methods for plasma control and disruption prediction, to surrogates for diagnostics. \cite{kwak_bayesian_2024,bannmann_bayesian_2024} used Bayesian modeling to improve plasma diagnostics, and \cite{szucs_detecting_2021,zarzoso_novel_2024} built ML models for detect plasma activity in experiments.

\section{Data for Numerical Experiments}
This section section contains additional data relevant to the numerical experiments. \Cref{table:out_of_sample_conditions} contains the sampling conditions, $\yb$, used to generate the samples in \cref{fig:contours_in_sample} and \cref{fig:out_of_sample_boxplot}.
\begin{table}[!h]
\centering
\footnotesize
\begin{tabular}{@{}lllllllrlllllllr@{}}
\toprule
&\multicolumn{6}{c}{In-sample conditions} &\multicolumn{8}{c}{Out-of-sample conditions}\\
\cmidrule(r){2-8} \cmidrule(r){9-16}
$\nfp$ & 2 & 3 & 4 & 5 & 6 & 7 & 8 & 2 & 3 & 3& 4 & 5 & 6 & 7 & 8 \\
$N$ & 0 & 1 & 1 & 1 & 1 & 1 & 1 & 0 & 0 & 1 & 1 & 1 & 1 & 1 & 1  \\
$A$ & 4.0 & 8.0 & 8.0 & 8.0 & 8.0 & 12.0 & 12.0 & 4.5 & 18.5 & 9.0 & 11.0 & 17.0 & 14.0 & 11.0 & 22.0 \\
$\meaniota$ & 0.3 & 1.3 & 1.6 & 1.6 & 2.6 & 3.0 & 3.5 & 0.36 & 0.5 & 0.5 & 1.4 & 2.5 & 2.0 & 3.7 & 3.5 \\
\bottomrule
\end{tabular}
\vspace{6pt}
\caption{In-sample conditions and out-of-sample conditions, $\yb$, used for generating stellarators in \cref{fig:contours_in_sample,fig:out_of_sample_boxplot}.}
\label{table:out_of_sample_conditions}
\end{table}

\end{document}